\documentclass[lettersize,journal]{IEEEtran}
\usepackage{cite}
\usepackage{amsmath,amssymb,amsfonts}
\usepackage{booktabs}
\usepackage{algorithmic}
\usepackage{graphicx}
\usepackage{textcomp}
\usepackage{orcidlink}
\usepackage{xcolor}
\usepackage{tikz}
\usepackage{etoolbox} 
\usepackage{listofitems} 
\usetikzlibrary{positioning, arrows.meta, shapes.geometric}
\usepackage{caption}
\usepackage{subcaption}
\usepackage{booktabs}

\def\BibTeX{{\rm B\kern-.05em{\sc i\kern-.025em b}\kern-.08em
    T\kern-.1667em\lower.7ex\hbox{E}\kern-.125emX}}
\begin{document}

\title{Artificial Intelligence-Driven Network-on-Chip Design Space Exploration: Neural Network Architectures for Design\\

}

\author{\IEEEauthorblockN {Amogh Anshu N} ,
\IEEEauthorblockA{\textit{Dept. of Electronics and Communication Engineering} 
\textit{UVCE},
Bangalore, India 
\orcidlink{0009-0003-8802-6191}}\\
\and
\IEEEauthorblockN{B P Harish},
\IEEEauthorblockA{\textit{Dept. of Electronics and Communication Engineering} 
\textit{UVCE},
Bangalore, India 
\orcidlink{0009-0007-4464-8406}
}
}

\maketitle

\begin{abstract}
Network-on-Chip (NoC) design requires exploring a high-dimensional configuration space to satisfy stringent throughput requirements and latency constraints. Traditional design space exploration techniques are often slow and struggle to handle complex, non-linear parameter interactions.  This work presents  a machine learning-driven framework that automates NoC design space exploration using BookSim simulations and reverse neural network models. Specifically, we compare three architectures—a Multi-Layer Perceptron (MLP), a Conditional Diffusion Model, and a Conditional Variational Autoencoder (CVAE)—to predict optimal NoC parameters given target performance metrics. Our pipeline generates over 150,000 simulation data points across varied mesh topologies. The Conditional Diffusion Model achieved the highest predictive accuracy, attaining a mean squared error (MSE) of 0.463 on unseen data. Furthermore, the proposed framework reduces design exploration time by several orders of magnitude, making it a practical solution for rapid and scalable NoC co-design.
\end{abstract}


\section{Introduction}
\IEEEPARstart{W}{ith} the rise of multi-core and many-core processors, Network-on-Chip (NoC) architectures have become the backbone of on-chip communication systems \cite{b1}. However, designing an efficient NoC remains a major challenge due to the vast number of configuration parameters and their non-trivial interactions. Designers must often trade off latency, throughput, and power while designing routing algorithms, buffer sizes, virtual channels, for given traffic injection rates \cite{b2}, \cite{b3}.

\noindent Traditional NoC design methodologies rely heavily on brute-force simulations or heuristic algorithms, which are computationally expensive and slow to converge, especially for large design spaces \cite{b4}. Furthermore, these approaches typically operate in the forward direction—predicting performance from given parameters—but this fails to address a more pressing inverse problem: given desired performance, what are the optimal parameters?

\noindent Recent advances in machine learning (ML) offer new tools for automating hardware design \cite{b5}, \cite{b6}. However, most prior work in ML-for-NoC focuses on forward modeling or classification tasks with little attention given to design. [Moreover, few studies compare how different ML models handle the many-to-one mapping that arises in reverse prediction.

\section{Related Work}
An extensive literature survey has been carried out and the state-of-art is presented under the following heads.
\subsection{Network-on-Chip Design Space Exploration}
\noindent Traditional NoC design space exploration has relied primarily on simulation-based approaches combined with optimization heuristics. Early work on NoC topology optimization using genetic algorithms,\cite{b7}, while subsequent research has explored various metaheuristic approaches including particle swarm optimization \cite{b8} and simulated annealing \cite{b9}.

\noindent Several analytical models have been proposed to reduce simulation overhead. Queueing theory-based models for NoC performance prediction  \cite{b10}, and support vector regression-based models for latency modeling for NoC\cite{b11} have been presented. However, these analytical approaches often sacrifice accuracy for speed, particularly when modeling complex traffic patterns and buffer dynamics.

\subsection{Machine Learning in Hardware Design}
\noindent The application of machine learning to hardware design optimization has gained significant traction in recent years. Neural network-based performance prediction for processor architectures have been developed \cite{b12}.
\noindent Earlier literature suggests use of support vector machines for NoC optimization \cite{b13}, focusing on forward prediction models. More recently, adaptive deep reinforcement learning has been applied to NoC routing algorithm design\cite{b14}, that failed to address the reverse parameter prediction problem.

\subsection{Generative Models for Design Optimization}
\noindent Generative models have shown promise in various design domains. Conditional Variational Autoencoders (CVAEs) have been successfully applied to circuit design \cite{b15}. Diffusion models, while newer, have demonstrated superior generation quality in image synthesis \cite{b16} and have begun finding applications in engineering design.  However, no prior work has systematically compared these generative approaches for NoC parameter optimization, representing a significant gap that the proposed work addresses. \\

\noindent  The contributions of the proposed work is summarized as follows:
\subsubsection{ Automated Simulation Framework}
This paper presents a Python-based framework for large-scale BookSim simulations with efficient configuration generation, parallel execution, and performance parsing.
\subsubsection{Reverse Parameter Prediction}
This paper reformulates NoC optimization as a supervised learning problem that predicts parameters from target latency and throughput specifications.

\subsubsection{ Architecture Comparison}
This paper implements and compares three distinct neural network architectures—MLP, Conditional Diffusion Model, and CVAE—providing insights into their relative strengths for NoC design optimization.
 
\subsubsection{Comprehensive Evaluation}
 This paper demonstrates a framework that achieves fast and accurate reverse predictions, reducing manual exploration and paving the way for AI-guided NoC co-design.

\noindent The remainder of this paper is organized as follows: Section III describes the methodology and neural network architectures. While Section IV presents the results and analysis, Section V discusses implications and limitations.  Section VI concludes with a roadmap for future work.

\section{Methodology}
\subsection{Problem Formulation}
\noindent This paper formulates the Network-on-Chip (NoC) design optimization as a reverse prediction problem. Given target performance specifications $\mathbf{P} = [\text{latency}_{\text{target}},\ \text{throughput}_{\text{target}}]$, we seek to predict optimal configuration parameters $\mathbf{X} = [\text{num\_vcs},\ \text{vc\_buf\_size},\ \text{injection\_rate},\ \text{packet\_size}]$ that achieve the desired performance. 
The variable \text{num\_vcs} refer to the number of virtual channels while \text{vc\_buf\_size} refers to the buffer size, \text{injection\_rate} refers to the packet injection rate and \text{packet\_size} refers to packet size.

\noindent Formally, we learn a mapping function $f: \mathbf{P} \rightarrow \mathbf{X}$ such that when $\mathbf{X}$ is used in a NoC simulation, the resulting performance $\mathbf{P}'$ minimizes the distance $\|\mathbf{P} - \mathbf{P}'\|$.

\subsection{Automated Simulation Framework}
\noindent This paper provides a framework that consists of four main components: \\
1. Configuration Generation: Systematic generation of BookSim configuration files with parameter combinations sampled from predefined ranges. \\
2. Parallel Execution: Multi-process simulation execution using Python's joblib library, with each process handling unique configuration files to avoid conflicts.\\
3. Output Parsing: Robust extraction of performance metrics from BookSim output using regular expressions, handling various output formats and error conditions.\\
4. Data Management: Structured storage of simulation results in pandas DataFrames with comprehensive metadata.\\

\begin{figure}
  \centering
  \includegraphics[width=0.34\textwidth]{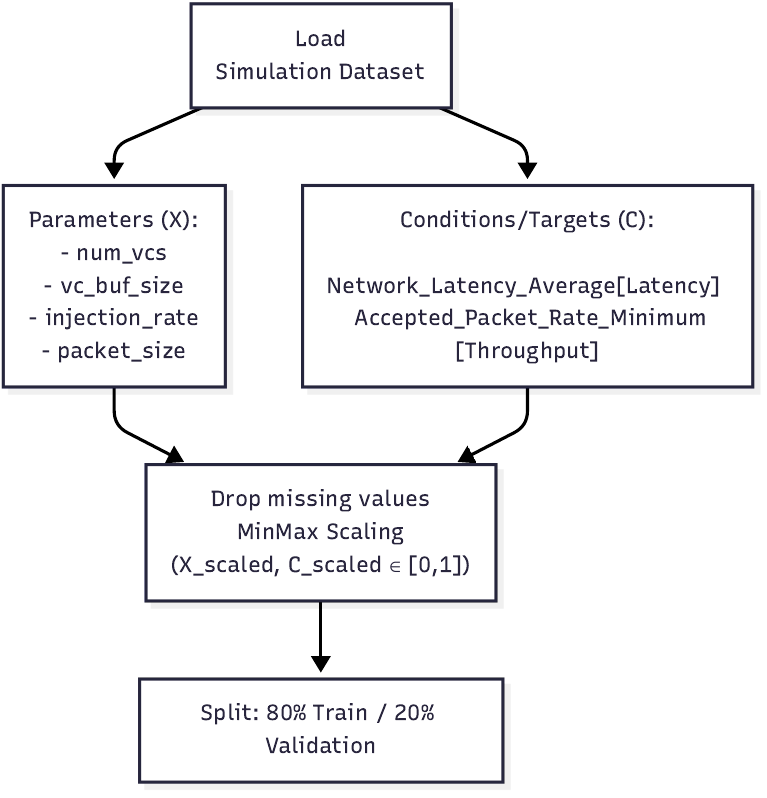}
  
\caption{Data preparation workflow: BookSim simulation outputs are cleaned, scaled, and split for supervised training}
 \label{fig:dprep}
\end{figure}

\noindent The framework handles over \textbf{150000} parameter combinations of 4X4 mesh topologies with dor routing function of BookSim 2.0 and uniform traffic across the following ranges:\\

\begin{itemize}
    \item {Number of Virtual Channels (num\_vcs)}: 2--20
    \item {Buffer Size (vc\_buf\_size)}: 2--10
    \item {Injection Rate}: 0.001--0.1
    \item {Packet Size}: 2--10 flits
\end{itemize}
\noindent Fig~\ref{fig:dprep} shows the data processing done post generation of training dataset obtained by multiple BookSim simulations

\subsection{Neural Network Architectures}

\textbf{1) Multi-Layer Perceptron (MLP)} \\
\noindent The baseline approach as shown in Fig~\ref{fig:mlp} uses a simple feedforward neural network to directly learn the mapping from performance targets to NoC parameters. The architecture consists of:

\begin{center}

\textbf{Input Layer (2 nodes)} $\rightarrow$ \textbf{Hidden Layer (64, ReLU)} 
$\rightarrow$ \textbf{Hidden Layer (64, ReLU)} $\rightarrow$ \textbf{Output Layer (4 nodes)}
\end{center}
\noindent where ReLU refers to the Rectified Linear Unit activation function, defined as $\text{ReLU}(x) = \max(0, x)$.
\begin{figure}

\begin{tikzpicture}[x=1.4cm, y=1.2cm, >=stealth, font=\small]

\tikzset{
  neuron/.style={circle, draw, minimum size=0.85cm},
  neuron missing/.style={
    draw=none, 
    scale=1.5,
    text height=0.333cm,
    execute at begin node=\color{black}$\vdots$
  }
}

\node[neuron] (input1) at (0,0.6) {};
\node[neuron] (input2) at (0,-0.6) {};

\node[neuron] (h1_1) at (2,1.2) {};
\node[neuron] (h1_2) at (2,0.4) {};
\node[neuron missing] at (2,-0.4) {};
\node[neuron] (h1_3) at (2,-1.2) {};

\node[neuron] (h2_1) at (4,1.2) {};
\node[neuron] (h2_2) at (4,0.4) {};
\node[neuron missing] at (4,-0.4) {};
\node[neuron] (h2_3) at (4,-1.2) {};

\node[neuron] (out1) at (6,1.2) {};
\node[neuron] (out2) at (6,0.4) {};
\node[neuron missing] at (6,-0.4) {};
\node[neuron] (out3) at (6,-1.2) {};

\node[align=center, above] at (0,1.6) {Input\\(2D)};
\node[align=center, above] at (2,1.6) {Hidden Layer 1\\(64D + ReLU)};
\node[align=center, above] at (4,1.6) {Hidden Layer 2\\(64D + ReLU)};
\node[align=center, above] at (6,1.6) {Output\\(4D)};

\foreach \i in {1,2} {
  \foreach \j in {1,2,3} {
    \draw[->] (input\i) -- (h1_\j);
  }
}

\foreach \i in {1,2,3} {
  \foreach \j in {1,2,3} {
    \draw[->] (h1_\i) -- (h2_\j);
  }
}

\foreach \i in {1,2,3} {
  \foreach \j in {1,2,3} {
    \draw[->] (h2_\i) -- (out\j);
  }
}

\end{tikzpicture}

\caption{MLP architecture for reverse NoC parameter prediction. The network maps 2D performance inputs to 4D configuration outputs.}
\label{fig:mlp}
\end{figure}

\noindent Let $\mathbf{y} \in \mathbb{R}^2$ denote the input vector representing the performance targets (i.e. latency and throughput), and $\hat{\mathbf{x}} \in \mathbb{R}^4$ be the predicted NoC parameter vector. The MLP learns a function $f_\theta: \mathbb{R}^2 \rightarrow \mathbb{R}^4$ parameterized by $\theta$  such that:

 \begin{equation}
    \hat{\mathbf{x}} = f_\theta(\mathbf{y}) = W_2 \, \sigma(W_1 \, \sigma(W_0 \mathbf{y} + \mathbf{b}_0) + \mathbf{b}_1) + \mathbf{b}_2
\end{equation}

\noindent where $W_i$ and $\mathbf{b}_i$ are the weight matrices and biases for layer $i$,  $\sigma(\cdot)$ is the ReLU activation function and $\theta$  is the set of all learnable parameters

\noindent The model is trained to minimize the mean squared error (MSE) between the predicted and true parameters:

\begin{equation}
    \mathcal{L}_{\text{MSE}} = \frac{1}{N} \sum_{i=1}^{N} \left\| \hat{\mathbf{x}}^{(i)} - \mathbf{x}^{(i)} \right\|^2
\end{equation}

\noindent  where $\mathbf{x}^{(i)}$ is the ground-truth parameter vector for the $i$-th sample in a dataset of size $N$.
\vspace{1em}

\vspace{1em}
\textbf{2) Conditional Variational Autoencoder (CVAE)} \\
\begin{figure}[ht]
  \centering
  \includegraphics[width=0.34\textwidth]{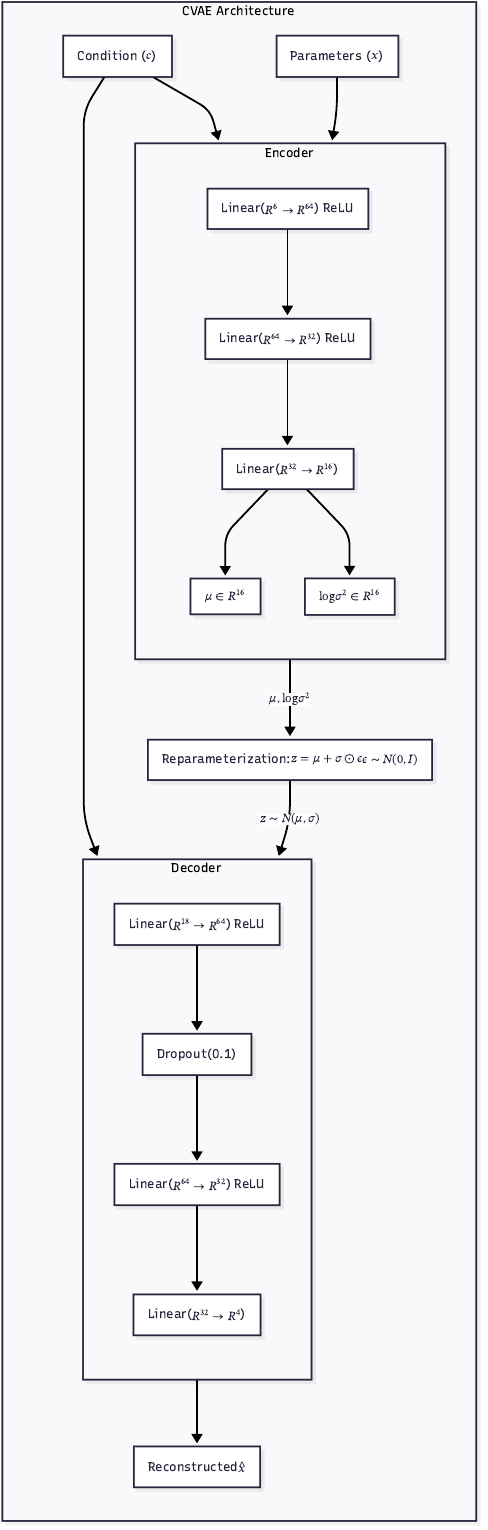}
  \caption{Architecture of Conditional Variational Autoencoder (CVAE) with encoder-decoder design.}
  \label{fig:arch}
\end{figure}

\noindent  The CVAE \cite{b17} learns a latent representation of the parameter space conditioned on performance targets such as latency and throughput. Fig~\ref{fig:arch} shows the architecture of the proposed CVAE model. \\
Let $\mathbf{x} \in \mathbb{R}^d$ denote a NoC parameter vector, and $\mathbf{y} \in \mathbb{R}^2$ the corresponding performance targets. The goal is to model the conditional distribution $p(\mathbf{x} \mid \mathbf{y})$.

\begin{itemize}
    \item \textbf{Encoder}: Maps the concatenated vector $[\mathbf{x}, \mathbf{y}]$ to a latent Gaussian distribution:
    \begin{align}
        q_\phi(\mathbf{z} \mid \mathbf{x}, \mathbf{y}) = \mathcal{N}\big(&\mathbf{z} \mid \boldsymbol{\mu}_\phi(\mathbf{x}, \mathbf{y}), \nonumber \\
        &\operatorname{diag}(\boldsymbol{\sigma}^2_\phi(\mathbf{x}, \mathbf{y}))\big)
    \end{align}
    where $\boldsymbol{\mu}_\phi$ and $\boldsymbol{\sigma}_\phi$ are neural networks parameterized by $\phi$ where $\phi$  is the set of learnable parameters.

    \item \textbf{Reparameterization}: A sample $\mathbf{z}$ is obtained via:
    \begin{equation}
        \mathbf{z} = \boldsymbol{\mu}_\phi + \boldsymbol{\sigma}_\phi \odot \boldsymbol{\epsilon}, \quad \boldsymbol{\epsilon} \sim \mathcal{N}(0, \mathbf{I})
    \end{equation}

    \item \textbf{Decoder}: Given the sampled latent code $\mathbf{z}$ and condition $\mathbf{y}$, reconstructs the parameters:
    \begin{equation}
        p_\theta(\mathbf{x} \mid \mathbf{z}, \mathbf{y}) = \mathcal{N}\left(\mathbf{x} \mid \hat{\mathbf{x}}_\theta(\mathbf{z}, \mathbf{y}), \sigma^2 \mathbf{I}\right)
    \end{equation}
    where $\hat{\mathbf{x}}_\theta$ is the decoder neural network.

    \item \textbf{Loss Function}: The objective combines a reconstruction term and a KL divergence regularization:
    \begin{align}
        \mathcal{L}(\theta, \phi) = \mathbb{E}_{q_\phi(\mathbf{z} \mid \mathbf{x}, \mathbf{y})} \left[ \| \mathbf{x} - \hat{\mathbf{x}}_\theta(\mathbf{z}, \mathbf{y}) \|^2 \right] \nonumber \\
        + \beta \cdot \mathrm{KL} \left( q_\phi(\mathbf{z} \mid \mathbf{x}, \mathbf{y}) \, \| \, \mathcal{N}(0, \mathbf{I}) \right)
    \end{align}
    where $\beta$ is a KL weight annealed during training to stabilize learning.
\end{itemize}

\textbf{3) Conditional Diffusion Model} \\

\begin{figure}
  \centering
  \includegraphics[width=0.45\textwidth]{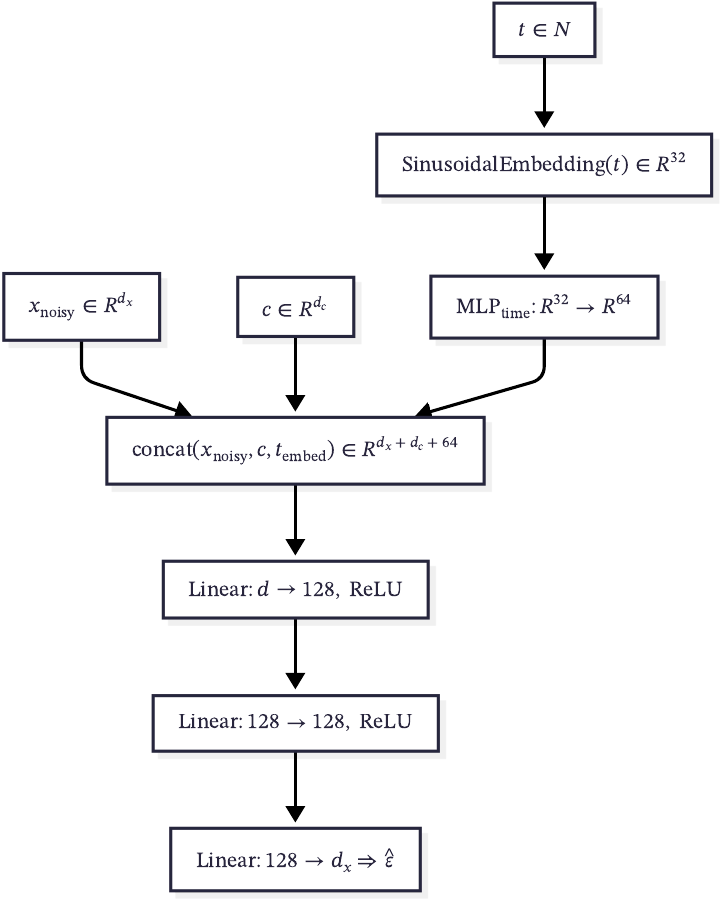}
  \caption{Conditional diffusion model: a denoising neural network learns to reverse noisy parameter vectors conditioned on performance metrics.}
  \label{fig:denoiser}
\end{figure}

\begin{figure}[ht]
  \centering
  \includegraphics[width=0.45\textwidth]{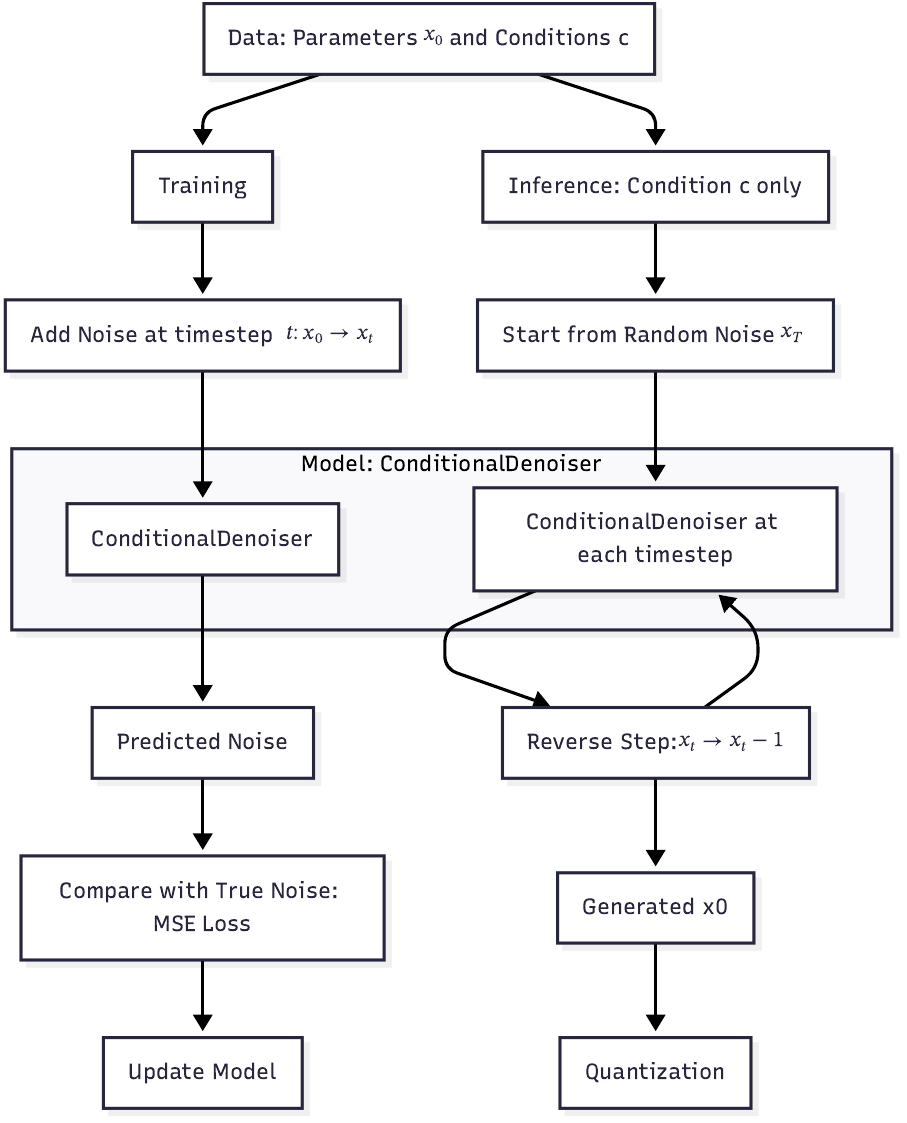}
  \caption{End-to-end training and inference flow of the conditional diffusion model with noise schedule and iterative denoising.}
  \label{fig:model_diff}
\end{figure}

\noindent   A denoising diffusion model's \cite{b16} conditional parameter generation is applied  as shown in Fig ~\ref{fig:denoiser}. The model learns to denoise parameter configurations conditioned on target performance:

\begin{itemize}
    \item \textbf{Forward Process}: Gradually adds Gaussian noise to parameter vectors over $T = 1000$ timesteps.
    \item \textbf{Reverse Process}: A neural network learns to predict and remove noise conditioned on performance targets.
    \item \textbf{Architecture}: Time embeddings combined with condition vectors, processed through multilayer perceptrons (MLPs). \\
     Fig ~\ref{fig:model_diff} shows the end-to-end training and inference flow of the conditional diffusion model.
\end{itemize}

\noindent  Mathematically, let $\mathbf{x}_0 \in \mathbb{R}^d$ denote a clean NoC parameter vector, and $\mathbf{y} \in \mathbb{R}^2$ the target performance condition (e.g., latency, throughput). The forward process corrupts $\mathbf{x}_0$ with Gaussian noise as:

\begin{equation}
    \mathbf{x}_t = \sqrt{\bar{\alpha}_t} \, \mathbf{x}_0 + \sqrt{1 - \bar{\alpha}_t} \, \boldsymbol{\epsilon}, \quad \boldsymbol{\epsilon} \sim \mathcal{N}(0, \mathbf{I})
\end{equation}

where $\alpha_t = 1 - \beta_t$ and $\bar{\alpha}_t = \prod_{s=1}^{t} \alpha_s$ is the cumulative product of the noise scales.

\noindent  A neural network $\boldsymbol{\epsilon}_\theta(\mathbf{x}_t, \mathbf{y}, t)$ is trained to predict the noise $\boldsymbol{\epsilon}$ via the loss:

\begin{equation}
    \mathcal{L}_{\text{MSE}} = \mathbb{E}_{\mathbf{x}_0, \mathbf{y}, \boldsymbol{\epsilon}, t} \left[ \left\| \boldsymbol{\epsilon} - \boldsymbol{\epsilon}_\theta(\mathbf{x}_t, \mathbf{y}, t) \right\|^2 \right]
\end{equation}

\noindent  During generation, the reverse process recursively denoises samples from a Gaussian prior using:

\begin{align}
    \mathbf{x}_{t-1} = \frac{1}{\sqrt{\alpha_t}} \Big( \mathbf{x}_t - \frac{1 - \alpha_t}{\sqrt{1 - \bar{\alpha}_t}} \, 
    \boldsymbol{\epsilon}_\theta(\mathbf{x}_t, \mathbf{y}, t) \Big) \nonumber \\
    + \sqrt{\beta_t} \, \mathbf{z}, \quad \mathbf{z} \sim \mathcal{N}(0, \mathbf{I})
\end{align}

\noindent  Sinusoidal time embeddings $\gamma(t) \in \mathbb{R}^{d_t}$ are used to encode timestep information and are concatenated with $\mathbf{x}_t$ and $\mathbf{y}$ before input to the denoising network.

\subsection{Training and Evaluation}
All models are trained using the Adam optimizer with learning rates tuned to individual architectures. We employ a 80\% - 20\% train - validation split and monitor convergence through validation loss. For discrete parameters (\texttt{num\_vcs}, \texttt{vc\_buf\_size}, \texttt{packet\_size}), we apply post-processing quantization / clamping to ensure valid configurations.

\section{Results and Analysis}

\subsection{Model Overview}

\noindent  Table~\ref{tab:architecture_comparison} compares the architectural and operational features of the three models evaluated. The MLP acts as a deterministic baseline, while CVAE and diffusion are generative models capable of producing multiple parameter configurations for the same performance target. This ability is critical in reverse modeling where many-to-one mappings naturally arise due to hardware design degeneracy.

\begin{table}[h!]
\centering
\caption{Model Architecture Comparison}
\label{tab:architecture_comparison}
\resizebox{\columnwidth}{!}{%
\begin{tabular}{lccc}
\toprule
\textbf{Feature} & \textbf{MLP}  & \textbf{CVAE} & \textbf{Diffusion} \\
\midrule
Architecture & Feedforward MLP & Conditional VAE  & Conditional DDPM \\
Output Type & Single prediction  & Multiple samples & Multiple samples \\
Probabilistic & No  & Yes & Yes \\
Discrete Param Handling & Post-hoc clamping  & Post-hoc clamping & Post-hoc quantization\\
Design Space Coverage & Low  & Moderate & High\\
Sampling Method & --  & Decoder with latent $z$ & Iterative denoising \\
\bottomrule
\end{tabular}
}
\label{tab:architecture_comparison}
\end{table}

\subsection{Training and Validation Performance}

\noindent  All models were trained using MSE loss on parameter values, with results shown in Table~\ref{tab:final_epoch_metrics}. The MLP achieves the lowest loss due to its direct mapping and absence of sampling variance. The CVAE exhibits comparable performance, though minor KL divergence instability was observed. The diffusion model shows slightly higher loss due to added noise during training but provides more flexible sampling at inference time.

\begin{table}[h]
\centering
\caption{Final Epoch Metrics of Training Loop}
\label{tab:final_epoch_metrics}
\resizebox{\columnwidth}{!}{%
\begin{tabular}{lccc}
\toprule
\textbf{Metric} & \textbf{MLP}  & \textbf{CVAE} & \textbf{Diffusion}\\
\midrule
Training Loss & 0.0466 & 0.0470 & 0.0651 (noise) \\
Validation Loss & 0.0467  & 0.0471 & 0.0664 (noise) \\
\bottomrule
\end{tabular}
}
\label{tab:final_epoch_metrics}
\end{table}

\noindent  While these metrics reflect learning performance, in order to fully capture the real-world usefulness of the predicted parameters, BookSim-based simulations are considered.

\subsection{BookSim-Based Evaluation of Reverse Models}

\noindent  To evaluate the functional accuracy of the predicted NoC parameters, we simulate each model’s output using BookSim and compare the resulting latency and throughput against the original target. Generative models (CVAE and diffusion) generate 10 samples per target, and the best-performing configuration is selected based on simulation error.

\begin{table}[ht]
\centering
\caption{BookSim Evaluation of Reverse Models (100 Samples)}
\label{tab:booksim_results}
\label{tab:booksim_results}
\begin{tabular}{lccc}
\toprule
\textbf{Model} & \textbf{MSE (Latency)} & \textbf{MSE (Throughput)} & \textbf{Total MSE} \\
\midrule
MLP        & 2.824043 & 0.000002 & 1.412023 \\
CVAE       & 10.288263 & 0.000004 & 5.144134 \\
Diffusion  & \textbf{0.926223} & 0.000005 & \textbf{0.463114} \\
\bottomrule
\end{tabular}
\end{table}

\noindent  As shown in Table~\ref{tab:booksim_results}, the diffusion model achieves the lowest average MSE across both latency and throughput. It outperforms both MLP and CVAE by effectively exploring diverse configurations and selecting the most accurate match. The MLP, though computationally efficient, is constrained by its single prediction and performs poorly in scenarios with multiple valid designs. The CVAE, while generative, struggles with variance and conditioning, often producing suboptimal parameter sets.

\subsection{Discussion}

\begin{itemize}
    \item \textbf{MLP} offers speed and simplicity but lacks flexibility in handling multiple valid outputs per target.
    \item \textbf{CVAE} demonstrates the value of sampling but suffers from high error variance and latent space instability.
    \item \textbf{Diffusion} balances diversity and accuracy, enabling it to better handle the many-to-one nature of the reverse mapping.
\end{itemize}

\noindent  These results highlight the limitations of deterministic models in reverse prediction and demonstrate the potential of generative architectures in capturing design degeneracy. However, the reliance on post-hoc simulation for validation points to a key area for improvement, discussed further in Section VI.

\section{Implications and Limitations}

\subsection{Implications}
\begin{itemize}
    \item \textbf{Reverse design automation:} This work demonstrates the feasibility of using AI models to infer NoC configuration parameters directly from target performance metrics, enabling faster and automated early-stage design exploration.

    \item \textbf{Performance-aware evaluation:} The integration of BookSim-based simulation into the evaluation process allows post-hoc validation of generated configurations, bridging the gap between model predictions and real-world NoC performance in terms of latency and throughput.

    \item \textbf{Support for design diversity:} Generative models such as the CVAE and diffusion models enable sampling of multiple valid configurations for the same performance target, which is critical in the presence of many-to-one mappings.

    \item \textbf{Differentiable optimization:} This work lays the foundation for incorporating differentiable surrogates or black-box optimization into training, enabling end-to-end fine-tuning with respect to actual performance metrics.
\end{itemize}

\vspace{1em}
\subsection{Limitations}
\begin{itemize}
    \item \textbf{Post-hoc performance evaluation:} The models are trained using MSE loss on parameter values, while performance metrics are only evaluated post-training via BookSim. This introduces a mismatch between the training objective and deployment criteria.

    \item \textbf{Non-differentiable simulation engine:} BookSim is non-differentiable and cannot be used for gradient-based optimization. This limits its integration in end-to-end training unless approximated by a differentiable surrogate model.

    \item \textbf{Handling of discrete variables:} NoC parameters such as buffer size and number of virtual channels are discrete by nature. Rounding and clamping which are used during inference can introduce errors in a best-case scenario or invalid configurations in a worst-case scenario.

    \item \textbf{Underperformance of CVAE:} Despite its generative capability, the CVAE model underperformed in simulation-based evaluation, likely due to posterior collapse or insufficient latent conditioning, warranting future investigation.

    \item \textbf{Limited evaluation metrics:} The current evaluation focuses solely on latency and throughput. Other critical design metrics such as power, area, and thermal constraints are not considered, limiting the framework’s applicability in holistic NoC co-design.
\end{itemize}

\section{Future Work}
\noindent  While the current framework evaluates the quality of reverse predictions using BookSim post hoc, future efforts can incorporate latency and throughput optimization directly into the training loop, either by using a BookSim surrogate model or by fine tuning with BookSim-in-the-loop.


\end{document}